\documentclass[10pt,twocolumn,letterpaper]{article}

\usepackage[pagenumbers]{cvpr}
\usepackage{graphicx}
\usepackage{amsmath}
\usepackage{amssymb}
\usepackage{booktabs}
\usepackage{color}
\usepackage{url}
\usepackage{multirow}
\usepackage{booktabs}
\usepackage{layouts}
\usepackage{adjustbox}
\usepackage[pagebackref,breaklinks,colorlinks]{hyperref}
\usepackage[capitalize]{cleveref}
\crefname{section}{Sec.}{Secs.}
\Crefname{section}{Section}{Sections}
\Crefname{table}{Table}{Tables}
\crefname{table}{Tab.}{Tabs.}

\usepackage{amsmath,amsfonts,bm}

\def\eqref#1{equation~\ref{#1}}

\def\1{\bm{1}}

\DeclareMathAlphabet{\mathsfit}{\encodingdefault}{\sfdefault}{m}{sl}
\SetMathAlphabet{\mathsfit}{bold}{\encodingdefault}{\sfdefault}{bx}{n}

\title{Continually Learning Self-Supervised Representations with Projected Functional Regularization}

\author{Alex Gomez-Villa\thanks{Computer Vision Center (CVC),  Universitat Autonoma of Barcelona (UAB), Barcelona, Spain}, Bartlomiej Twardowski\footnotemark[1],
Lu Yu\footnotemark[1],
 Andrew D. Bagdanov\thanks{ Media Integration and
Communication Center, Florence, Italy},
 Joost van de Weijer\footnotemark[1]\\
{\tt\small \{agomezvi,btwardowski,luyu,joost\}@cvc.uab.es, andrew.bagdanov@unifi.it}
}

\iftrue
\newcommand{\JW}[1]{{\color{magenta}{\bf JW: #1}}}

\definecolor{othergreen}{rgb}{0.0, 0.5, 0.0}

\else
\newcommand{\JW}[1]{}
\fi

\newcommand{\minisection}[1]{\vspace{0.04in} \noindent {\bf #1}\ \ }

\begin{document}

\maketitle

\begin{abstract}
Recent self-supervised learning methods are able to learn high-quality image representations and are closing the gap with supervised approaches. However, these methods are unable to acquire new knowledge incrementally -- they are, in fact, mostly used only as a pre-training phase over IID data. In this work we investigate self-supervised methods in continual learning regimes without any replay mechanism.
We show that naive functional regularization, also known as feature distillation, leads to lower plasticity and limits continual learning performance. Instead, we propose Projected Functional Regularization in which a separate temporal projection network ensures that the newly learned feature space preserves information of the previous one, while at the same time allowing for the learning of new features. This prevents forgetting while maintaining the plasticity of the learner.
Comparison with other incremental learning approaches applied to self-supervision demonstrates that our method obtains competitive performance in different scenarios and on multiple datasets.
\end{abstract}

\section{Introduction}

Self-supervised learning aims to learn high-quality image representations without the need for human annotations. A recent set of works has shown that self-supervised learning can achieve performance close to that of supervised learning~\cite{chen2020SimCLR, chen2021SimSiam, NEURIPS2020BYOL, caron2020SwAV}, and that learned representations transferred to downstream tasks are sometimes even superior to fully-supervised representation learning~\cite{caron2021dino}. These methods learn representations that are invariant with respect to a set of data augmentations. They are typically trained with contrastive losses in which multiple views of the same image (computed by applying different data augmentations) are mapped close together, whereas representations of other images are mapped far away. However, several methods show that only encouraging similarity between views from the same image (without any explicit loss to promote the distancing of negative pairs) can also obtain excellent performance~\cite{chen2021SimSiam, NEURIPS2020BYOL}. These methods apply various mechanisms to prevent trivial solutions, including asymmetric architectures and the use of momentum updates of the model. 

Recent works on self-supervised learning have in common that they assume that all training data is available during the training process. However, in many real-world applications the learner must cope with non-stationary data in which they are exposed to tasks with varying distributions of data. Continual learning relaxes the IID assumption that underlies most learning methods and studies the design of algorithms that learn from data with shifting distributions. Naively training a learner on such data, for example by simply continuing stochastic gradient descent, leads to catastrophic forgetting \cite{mccloskey1989catastrophic}. A variety of approaches have been proposed including various types of regularization \cite{li2017learning,kirkpatrick2017overcoming, zenke2017continual, aljundi2018memory}, data replay~\cite{rebuffi2017icarl, castro2018end, wu2019large, hou2019learning}, pseudo replay~\cite{shin2017continual,wu2018memory}, and growing architectures~\cite{rusu2016progressive}. Even though there is some work on unsupervised continual learning~\cite{rusu2016progressive, fernando2017pathnet, mallya2018packnet, li2019learn}, the vast majority of existing work is on supervised continual learning~\cite{parisi2019continual,pfulb2019comprehensive}. 

Earlier works on self-supervised learning was based on pretext tasks like predicting rotation~\cite{gidaris2018_pretext_rotation}, determining patch position~\cite{doersch2015_pretext_patches}, or solving jigsaw puzzles in images~\cite{noroozi2016_pretext_puzzle}. Labels for these discriminative pretext tasks can be automatically computed and allow learning of meaningful feature representations of images. Recently, researchers are adapting contrastive methods for unlabeled data and operating more at an instance-level augmentation while looking for similarity or contrastive samples~\cite{chen2020SimCLR, NEURIPS2020BYOL, caron2020SwAV, zbontar2021barlow}. These methods rely heavily on stochastic data augmentation to produce enough similar examples to learn representations. Negative examples are randomly sampled or not used at all~\cite{chen2021SimSiam}. The results are impressive and are competitive with many supervised methods on downstream tasks~\cite{caron2021dino}.

We propose an approach to \emph{continual} self-supervised learning that is able to learn high-quality visual feature representations from non-IID data. The learner is exposed to a changing distribution and, while learning new features on current task data, should prevent forgetting of previously acquired knowledge. These representations should, at the end of training, be applicable to a wide range of downstream tasks.  We focus on the more restrictive, memory-free continual learning setting in which the learner is not allowed to store any samples from previous tasks. This scenario is realistic in many scenarios where data privacy and security is fundamental and often legislatively regulated. 

The main contributions of this work are twofold. First, we propose a new method, called \textit{projected functional regularization}, to alleviate forgetting during unsupervised representation learning without the need for an external memory of samples from previous tasks. This technique is an extension of Learning without Forgetting (LwF) and distillation in feature space. To improve the plasticity of the method we introduce a \emph{temporal projection network} that provides more freedom to learn features from the current task. Secondly, we propose a set of experiments over benchmark datasets to compare with other state-of-the-art methods and use different scenarios to evaluate the functional projection role in the context of continual self-supervised representation learning. We show that the additional projection to past tasks results in better representation learning during class incremental training sessions. Without any adjustment, evaluation on a truly class incremental scenario -- with only a single class per task, where many class incremental methods cannot be directly applied -- our method still prevents forgetting and is able to progressively learn new features. Furthermore, we confirm that our method is generic and the results are not restricted to a particular self-supervised learning approach. In a variety of experimental settings the transferability of the learned features to different downstream tasks is maintained, confirming that the network is incrementally learning more robust representations.
The influence of the regularization strength is analyzed for different regularization methods applied to self-supervised continual learning and results clearly shows the benefit of the proposed additional projector resulting both in improved plasticity (i.e. lower intransigence) and less forgetting.


\section{Related Work}\label{sec:Related_Work}
Both self-supervised and continual learning have gathered increasing interest in recent years. We briefly review the literature on both topics before articulating our contribution which combines elements of both in the form of continual self-supervised representation learning. 

\minisection{Self-supervised learning.}
Self-supervised learning has proved useful for many applications. In order to learn representations useful for a downstream task, a self-supervised pretext task can be introduced to avoid supervision. Many pretext tasks were investigated for learning image representations, including rotation prediction~\cite{gidaris2018_pretext_rotation}, solving jigsaw puzzles~\cite{noroozi2016_pretext_puzzle}, determining relative patch positions~\cite{doersch2015_pretext_patches}, predicting surrogate classes~\cite{dosovitskiy2014_pretext_surrogate}, and image colorization~\cite{zhang2016_pretext_colorization}).

In the last few years, the gap between supervised and self-supervised learning is being closed. This is primarily due to methods based on data augmentation and contrastive-like learning in which two samples are considered either similar or different to each other. This has links to earlier contrastive methods used in metric learning~\cite{hadsell2006dimensionality} and some extensions using triplet losses~\cite{weinberger2006triplet}. However, in the unsupervised setting without labels, different approaches must be used for creating such pairs. In SimCLR~\cite{chen2020SimCLR}, similar samples are created by augmenting an input image with a random distortion, while dissimilar ones are chosen by random. To make contrastive training more efficient, the MoCo method~\cite{he2020MOCO, chen2020MOCOv2} uses a memory bank for learned embeddings which enables efficient sampling. This memory is kept synchronized with the rest of the network during training by using a momentum encoder. The SwAV approach uses online clustering over the embedded samples~\cite{caron2020SwAV}. SwAV does not sample negative exemplars, however, other cluster prototypes can play the role of negative examples.

Interesting are methods without any explicit contrastive pairs. The BYOL approach proposed by~\cite{NEURIPS2020BYOL} is based on an asymmetric network with an additional MLP predictor between two outputs of the two branches. One branch is kept ``offline'' and updated by a momentum encoder. SimSiam~\cite{chen2021SimSiam} goes even further and offers a simplified solution without a momentum encoder and moreover works well without a very large mini-batch size. BarlowTwins is another simplified solution like SimSiam which uses a loss function based on correlations between each pair in a current training mini-batch~\cite{zbontar2021barlow}. Negatives are implicitly assumed to be available in each mini-batch. No asymmetry is used by the BarlowTwins network, but a larger embedding size and bigger mini-batches are preferred in this method in comparison to SimSiam.

\minisection{Continual learning.}
Existing continual learning methods can be broadly divided into replay-based, architecture-based, and regularization-based methods~\cite{delange2021continual, masana2020class}. Replay-based methods save a small amount of  data from previously seen tasks~\cite{bang2021rainbow, chaudhry2019tiny} or generate synthetic data with a generative model~\cite{wang2021ordisco, zhai2021hyper}. 
Architecture-based method activate different subsets of network parameters for different tasks by allowing model parameters to grow linearly with the number of tasks. Previous works following this strategy include DER~\cite{yan2021dynamically}, Piggyback~\cite{mallya2018piggyback}, PackNet~\cite{mallya2018packnet}, DAN~\cite{rosenfeld2018incremental}, HAT~\cite{serra2018overcoming}, Ternary Masks~\cite{masana2021ternary} and PathNet~\cite{fernando2017pathnet}. Regularization-based methods add an additional regularization term derived from knowledge of previous tasks to the training loss. This can be done by either regularizing the weight space (constraining important parameters)~\cite{shi2021continual,tang2021layerwise} or the functional space (constraining predictions or intermediate features)~\cite{douillard2020podnet, cheraghian2021semantic, hu2021distilling}. EWC~\cite{kirkpatrick2017overcoming}, MAS ~\cite{aljundi2018memory}, REWC~\cite{liu2018rotate}, SI~\cite{zenke2017continual}, and RWalk~\cite{chaudhry2018riemannian} constrain the importance of network parameters to prevent forgetting. Methods such as LwF~\cite{li2017learning}, LwM~\cite{dhar2019learning} and BiC~\cite{wu2019large} instead leverage knowledge distillation to regularize features or predictions.

Our approach, called \emph{Projected Functional Regularization} is a functional regularization approach. Normally, these approaches distill information at the class-prediction level between an old and new model. However, in self-supervised learning this has to be applied to the embedding output. Regularizing the embedding layer is known to undermine plasticity~\cite{douillard2020podnet}, we therefore propose an additional temporal projection network that maps between the latent spaces of current and previous model. We show that this regularization prevents forgetting while obtaining improved plasticity. 

\minisection{Continual representation learning.}
Continual unsupervised representation learning was investigated by~\cite{rao2019curl} with an approach based on variational autoencoders and a Gaussian mixture model. Still, detection of new clusters and model expansion is necessary. In a very recent paper, the authors used contrastive self-supervised learning with a memory buffer for storing exemplars~\cite{madaan2021rethinking}. They proposed the LUMP method in which images from the current task and previous tasks are combined with CutMix for continual training. One of the main differences between LUMP and our approach is that ours does not require storing any data from previous tasks.

Our contribution is fundamentally different from methods using self-supervised learning to improve the learning of a sequence of supervised tasks~\cite{guan2021reduce,zhu2021prototype}. Their objective is not to learn from unlabeled data, but rather to use self-supervised learning to further enrich the feature representation. The hypothesis of these works is that, for class incremental learning scenario, the features learned via self-supervision will be more generic than ones learned from task-bounded discrimination problems.

\section{Continual Self-supervised Representation Learning}\label{sec:Methods}

We begin with a discussion of self-supervised representation learning, and then describe our proposed Projected Functional Regularization (PFR) approach for continual learning of self-supervised representations without the need of any memory or replay.

\begin{figure}[t]
\begin{center}
\includegraphics[width=1\linewidth]{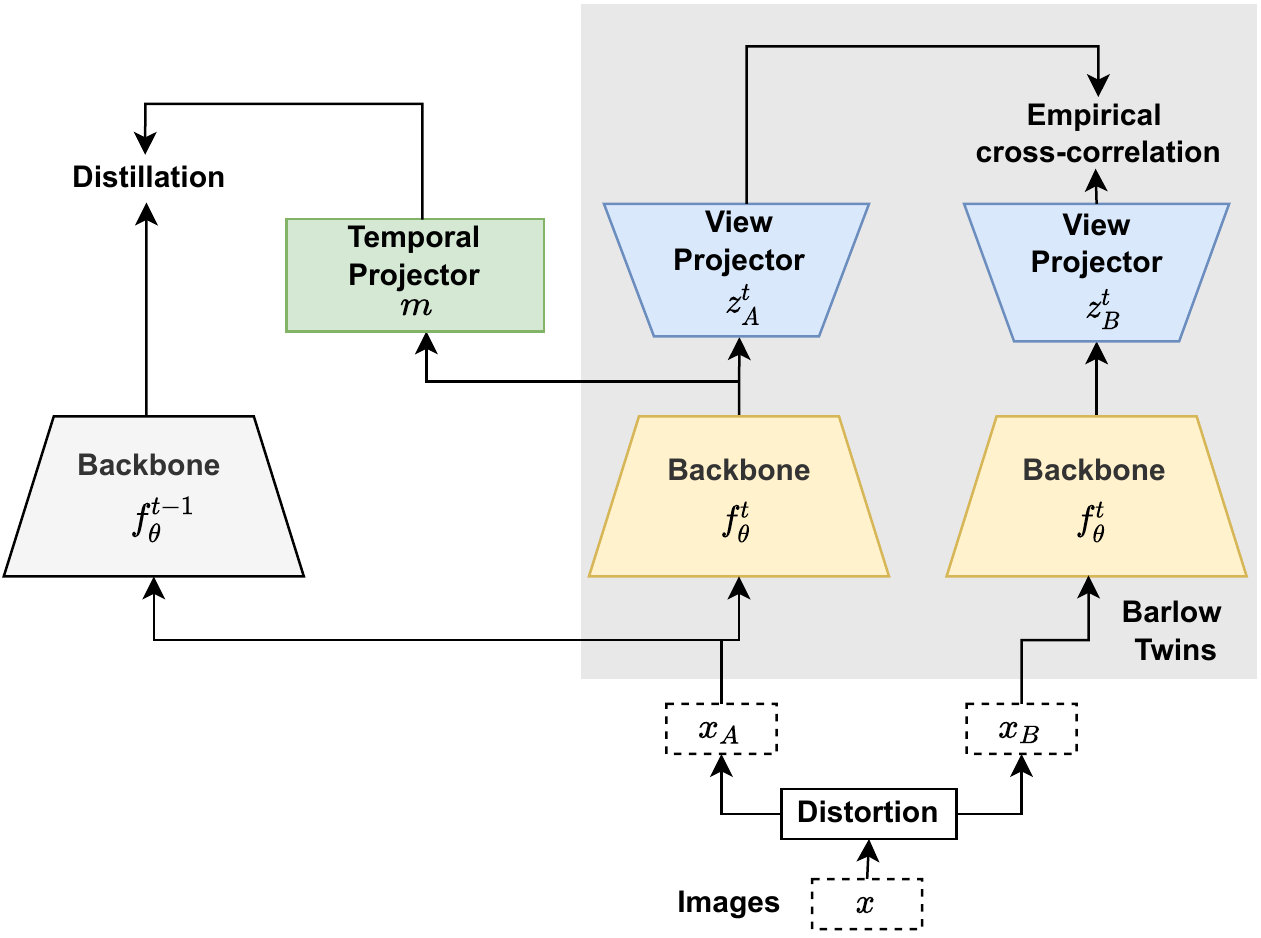}
\caption{Self-supervised continual learning with Projected Functional Regularization. Instead of performing feature distillation directly between the previous task backbone and the new one, we use a \emph{learned temporal projection} between the two feature spaces.}
\label{fig:method}
\end{center}
\end{figure}

\subsection{Self-supervised representation learning}
In recent works on self-supervised learning the aim is to learn a network 
$f_\theta:\mathcal{X}\rightarrow \mathcal{F}$ that maps from input space $\mathcal{X}$ to output feature representation space $\mathcal{F}$. This network is learned on unlabeled input data $x$ drawn from distribution $\mathcal{D}$. The aim is then to exploit the learned feature representation to perform any variety of downstream tasks. As an example, for the downstream task of classification on some target domain, we have training data $\mathcal{D}^t=\{x^t_i,y^t_i\}$ on which we learn a classifier $g_\phi:\mathcal{F}\rightarrow \mathcal{Y}$ (with $\mathcal{Y}$ being the output space) that minimizes a loss $\mathcal{L}=\ell \left(y^t,\hat y^t=g_{\phi}\left(f_{\theta}\left(x^t\right)\right)\right)$. Adaptation to the target domain might only optimize the weights $\phi$ while keeping $\theta$ fixed on the target data, or instead might also allow $\theta$ to be fine-tuned on the target data.

In this paper we apply the  BarlowTwins~\cite{zbontar2021barlow} approach to self-supervised learning of the representation network $f_{\theta}$. However, the proposed method is general and can be applied to other self-supervised methods. 
BarlowTwins does not require explicit negative samples and achieves competitive performance while remaining computationally efficient, assuming that negatives are available in each mini-batch to calculate correlation between all samples in it.
The BarlowTwins architecture has two branches (see the shaded area in Fig.~\ref{fig:method}). In both branches a projector network $z:\mathcal{F}\rightarrow \mathcal{Z}$ is used. For the sake of notational simplicity, we do not make explicit the parameters of the network $z$ since it is not used by downstream tasks. The parameters in the backbone and projector layer are shared between the branches.


The network is trained by minimizing an invariance and a redundancy reduction term in the loss function~\cite{zbontar2021barlow}. Here, different augmented views $X_A$ and $X_B$ of the same data samples $X$ are taken from the set of data augmentations $\mathcal{D^*}$. This leads to the loss defined as:
\begin{equation}
\mathcal{L}_c = \mathbb{E}_{X_A, X_B \sim \mathcal{D^*} } \!\! \left[ \sum_i  (1-\mathcal{C}_{ii})^2 + \lambda \sum_{i} \sum_{j \neq i} {\mathcal{C}_{ij}}^2 \right] \!\! ,
\label{eq:lossBarlow}
\end{equation}
where $\lambda$ is a positive constant trade-off parameter between both terms, and where $\mathcal{C}$ is the cross-correlation matrix computed between the representations $z$ of all samples $X_A$ and $X_B$ in a mini-batch indexed by $b$:
\begin{equation}
\mathcal{C}_{ij} = \frac{
\sum_b z^A_{b,i} z^B_{b,j}}
{\sqrt{\sum_b {(z^A_{b,i})}^2} \sqrt{\sum_b {(z^B_{b,j})}^2}},
\label{eq:crosscorr}
\end{equation}
The cross-correlation matrix $\mathcal{C}$ has values in the range of -1.0 (worst) to 1.0 (best) correlation between the projector's outputs: $Z_A = z(f_\theta(X_A))$ and $Z_B = z(f_\theta(X_B))$. The invariance term of the loss function encourages the diagonal elements to be 1. As such the learned embedding will be invariant to the applied data augmentations. At the same time, the second term (redundancy reduction) keeps the off-diagonal elements close to zero and decorrelates the outputs of non-related images.

\subsection{Projected Functional Regularization}
Current work on self-supervised learning considers the above scenario where the learner has access to a single, large dataset which can be revisited multiple times to learn the optimal feature extractor $f_{\theta}$. However, for many real-world scenarios this is an unrealistic setup and the learner will have to learn the optimal feature extractor $f_{\theta}$ from a stream of data drawn from a distribution that varies over time.

We consider scenarios in which the learner must learn from a set of tasks, each containing data drawn from a different distribution. We consider the tasks $T=\{1...c\}$ where $c$ is the current task and the data of task $t$ follows the distributions $\mathcal{D}_t$. In this case we would like to find the parameters $\theta$ of the feature extractor $f_{\theta}$ that minimize the summed loss over all tasks up to the current one $c$: 
\begin{equation}
    \arg\min_{\theta}\sum_{t=1}^c\mathcal{L}_c^t,
\end{equation}
where $\mathcal{L}_c^t=\mathbb{E}_{X_A, X_B \sim \mathcal{D}^*_t } [\mathcal{L}_c]$ and $\mathcal{L}_c$ is defined as in Eq.~\ref{eq:lossBarlow}.
Again, $\mathcal{D}_t^*$ refers to the set of augmented samples from $\mathcal{D}_t$ (i.e. the data from task $t$). However, during continual training we only have access to the data of one task, meaning that the optimal parameters must be found using only the current data $\mathcal{D}_c$. Naive fine-tuning results in parameters optimal for task $c$, however leads to catastrophic forgetting of knowledge acquired during previous tasks.

Regularization methods are among the most successful at addressing catastrophic forgetting, especially for scenarios where storing of any data from previous tasks is prohibited (which is the objective in this article). Regularization methods can be divided into two important groups: weight regularization approaches~\cite{kirkpatrick2017overcoming,zenke2017continual,aljundi2018memory}, which aim to find a set of weights that is both good for the current task while incurring only a small increase in loss on previous tasks, and functional regularization methods (also known as data regularization methods) which optimize weights for new tasks while incurring only minimal changes in the network outputs on previous tasks~\cite{Hou_2018_ECCV,titsias2020functional,pan2020continual}.

The canonical example of functional regularization, called Learning without Forgetting (LwF), was introduced in~\cite{Hou_2018_ECCV} and is based on knowledge distillation~\cite{hinton2014distilling}. It was proposed for supervised continual learning and introduces an additional loss that prevents the class predictions of previous tasks on the current data from undergoing large changes while training on the current task data. This loss cannot be directly applied to self-supervised learning since it requires class predictions. However, several continual learning works have extended this idea to feature layers by replacing the modified cross-entropy distillation loss with a distance (typically L1 or L2) which can be applied to any layer output~\cite{liu2020generative,douillard2020podnet,yu2020semantic}. We will refer to this as \emph{feature distillation} ($FD$) and it leads to the following loss when training task $t$: 
\begin{multline}
     \mathcal{L}_c^t+\lambda_{fd}\mathbb{E}_{x_a, x_b \sim \mathcal{D}^*_i }[\parallel f_{\theta^t}\left(x_a\right)-f_{\theta^{t-1}}\left(x_a\right) \parallel \\ +\parallel f_{\theta^t}\left(x_b\right)-f_{\theta^{t-1}}\left(x_b\right) \parallel],
     \label{eq:fd}
\end{multline}
where $\theta^{t-1}$ refers to the parameters learned after training up to task $t-1$, and $\lambda_{fd}$ defines the importance of the regularization term. 

The regularization imposed on class predictions in the original LwF paper~\cite{Hou_2018_ECCV} is not very restrictive: the weights can still significantly vary as long as the final network predictions do not significantly vary. It has been observed in the literature, however, that feature distillation is very restrictive and leads to continual learning methods with low plasticity~\cite{douillard2020podnet}. In addition, this loss directly penalizes the learning of new features since these would lead to a difference between the new and old model output $\parallel f_{\theta^t}\left(x\right)-f_{\theta^{t-1}}\left(x\right) \parallel$. To address this problem we propose \emph{Projected Functional Regularization} ($PFR$).

We would like the network to retain previous feature representation while allowing it to learn new features learned on new tasks. These new features should not be directly penalized by regularization. To do so, we introduce a \emph{temporal projection network} $m:\mathcal{Z}\rightarrow \mathcal{Z}$  that maps the embedding learned on the current task back to the embedding learned on the previous ones (see Figure~\ref{fig:method}). The new loss is:
\begin{multline}
     \mathcal{L}_c^t+\lambda_{pfr}\mathbb{E}_{x_a, x_b \sim \mathcal{D}^*_i }[\mathcal{S} ( m(f_{\theta^t}\left(x_a\right)), f_{\theta^{t-1}}\left(x_a\right) ) \\ +\mathcal{S} ( m(f_{\theta^t}\left(x_b\right)), f_{\theta^{t-1}}\left(x_b\right) )]
     \label{eq:pfr}
\end{multline}
where $S(\cdot, \cdot)$ is a cosine similarity:
\begin{equation}
\mathcal{S}(a,b) = - \frac{a^T b}{||a||_2 ||b||_2}.
\label{eq:dist_cosine}
\end{equation}

New features learned in $f_{\theta^t}\left(x\right)$ do not directly result in an increased loss as long as they lie in the null-space of $m$. As a consequence this loss prevents forgetting of information of previous tasks while maintaining plasticity to adapt to new tasks.

\section{Experimental Results}\label{sec:Results}

Here we report on a variety of experiments performed to evaluate the performance of Projected Functional Regularization for continual self-supervised representation learning without the need of an exemplar memory and replay.

\subsection{Datasets}

We use the following datasets in our evaluation:
\begin{itemize}
\item \textbf{CIFAR-100}: Proposed by~\cite{krizhevsky2009learning}, this dataset consists 100 object classes in 45,000 images for training, 5,000 for validation, and 10,000 for test with 100 classes. All images are 32$\times$32 pixels.
\item \textbf{Tiny ImageNet}: A rescaled subset of 200 ImageNet~\cite{deng2009imagenet} classes used in~\cite{tinyIM} and containing 64$\times$64 pixel images. Each class has 500 training images, 50 validation images and 50 test images.
\item \textbf{SVHN}: contains 32$\times$32 pixel images of from house numbers. There are 10 classes with 73,257 training images and 26,032 test images. From we split $5\%$ of the training images to use as a validation set.
\item \textbf{Cars}: Was introduced in ~\cite{krause20133d}. contains $16,185$ images of $196$ cars classes which includes 8,144 as train set and 8,041 as test set. 
\item \textbf{Aircraft}: Was proposed in\cite{maji2013fine} and consists 6,667 images for training and 3,333 for testing of 100 classes.
\end{itemize}
The last three datasets are used for evaluating our proposed method on downstream tasks. We downscale images to 64$\times$64 for Cars and Aircraft in our experiments.

\subsection{Training procedure and baseline methods}
\label{sec:train}

In all experiments, we train a ResNet-18~\cite{he2016deep} using SGD with an initial learning rate of $0.06$ and a weight decay of $0.0001$. The network is trained with cosine annealing for the first $1500$ epochs. After these epochs of cosine annealing, the learning rate is reduced by a factor of $0.8$ for the both projectors (view and temporal) and $0.4$ for the backbone. The data augmentation process is the same as in BarlowTwins (which was taken from SimCLR~\cite{chen2020SimCLR}). As a temporal projector, we use an MLP with a linear layer of 512 neurons followed by a batch normalization, Relu and a second linear layer of 256 neurons for CIFAR-100 and a linear layer with 512 neurons followed by a ReLu for TinyImageNet.

Downstream task classifiers are by default linear with a cross-entropy loss and are trained with Adam optimizer with a learning rate $5e$-$3$ for CIFAR-100 and $5e$-$2$ for TinyImageNet. We use validation data to implement a patience scheme that lowers the learning rate by a factor of $0.3$ up to three times while training a downstream task classifier.

In our experiments we compare with the following baseline methods: 
\begin{itemize}
\item \textbf{Fine-tuning (FT)}: The network is trained sequentially on each task without access to previous data and with no mitigation of catastrophic forgetting.
\item \textbf{Single Task}: We perform joint training with fine-tuning on all data which provides an upper bound. 
\item \textbf{Continual Joint Training (CJ)}: We continually perform joint training on the entire dataset seen so far. This is provides a tighter upper bound than Single Task~\cite{ash2020warm}.
\item \textbf{Feature Distillation (FD)}: Knowledge distillation is used as in LwF~\cite{li2017learning} to retain representation from previous tasks. We use the L2 distance as the regularization term, as is also proposed by other methods performing knowledge distillation on feature embeddings~\cite{liu2020generative, yu2020semantic, douillard2020podnet}.
\item \textbf{Elastic Weight Consolidation (EWC)}: We use the regularization method from~\cite{kirkpatrick2017overcoming} with a contrastive loss used to estimate the diagonal of the Fisher Information Matrix.
\end{itemize}
Note that we only compare to exemplar-free methods and exclude methods that require replay from our comparison. \footnote{code available at \url{https://github.com/alviur/CVPR_PFR.git}}

\subsection{Continual representation learning}


In this experiment we evaluate all methods in the incremental representation learning setting. The most straightforward way of doing this is to use the class incremental learning setting without access to labels. Specifically, we split datasets into four equal task as done in~\cite{rebuffi2017icarl}. In each task we learn a self-supervised representation and in the evaluation phase we train a linear classifier using the trained backbone encoder. In order to assess the learned representation, we use all available test data to obtain the overall task-agnostic performance evaluation\footnote{Note that we use \emph{task agnostic} in this paper to refer to the class-incremental learning evaluation~\cite{masana2020class}.}

In Table~\ref{tab:cifar100_main} we present the results for all methods. After the final task, the upper bound CJ obtains $63.6\%$, while a simple fine-tuning (FT) method reaches $54.8\%$. This is the gap where methods using regularization can improve. Joint training on all data at once outperforms CJ by $1.8\%$. PFR obtains an accuracy after the final task of $59.7\%$, while other regularization methods FD and EWC reach $57.8\%$ and $55.0\%$, respectively.

\begin{table}[]
\caption{ (top) Accuracy on CIFAR-100 with 4 tasks of 25 classes for incremental, self-supervised training. The learned representation is evaluated using a linear classifier over all classes after each task. (bottom) Evaluation of the same trained models on a different downstream task - classification using SVHN dataset. Mean and standard deviation over five runs are provided.}
\centering
\begin{adjustbox}{max width=\columnwidth}
\label{tab:cifar100_main}
\begin{tabular}{l|llll}
\midrule
\multicolumn{5}{c}{CIFAR100}  \\
\midrule
\textbf{Method}     & \textbf{Task 1}  & \textbf{Task 2 } & \textbf{Task 3}  & \textbf{Task 4} \\
\midrule\midrule
Single  &    -     &    -     & -  & 65.4$\pm$1.4      \\
CJ  & 53.3$\pm$0.7 & 58.4$\pm$0.8 & 60.8$\pm$1.1 & 63.6$\pm$1.5       \\
\midrule
FD  & 53.3$\pm$0.4 & 55.6$\pm$0.5 & 56.8$\pm$1.0 & 57.8$\pm$0.5      \\
EWC & 53.0$\pm$0.3 & 53.1$\pm$0.3 & 53.8$\pm$0.6 & 55.0$\pm$0.4     \\
PFR & 53.2$\pm$0.4 & 56.4$\pm$0.4 & 58.2$\pm$0.3 & \textbf{59.7}$\pm$0.3   \\
\midrule
FT  & 53.4$\pm$0.5 & 53.0$\pm$0.7 & 54.6$\pm$0.2 & 54.8$\pm$1.0  \\
\midrule
\multicolumn{5}{c}{SVHN}  \\
\midrule
\textbf{Method}      & \textbf{Task 1}  & \textbf{Task 2}  & \textbf{Task 3}  & \textbf{Task 4} \\
\midrule\midrule
Single  & - &  -  & -  & 64.0$\pm$1.4      \\
CJ  &60.6$\pm$2.6 &  63.2$\pm$1.7 &  63.4$\pm$1.3 & 63.0$\pm$1.3    \\
\midrule
FD  & 60.4$\pm$2.7 & 62.7$\pm$2.2 & 64.3$\pm$1.7 & 65.8$\pm$1.5     \\
EWC & 61.3$\pm$2.3 &  62.4$\pm$1.8 & 64.1$\pm$1.2 &  64.5$\pm$1.5  \\
PFR & 60.4$\pm$2.7 &  63.5$\pm$2.3 &  66.2$\pm$1.9 &  \textbf{68.0}$\pm$1.5 \\
\midrule
FT  & 60.4$\pm$2.7 &  61.0$\pm$1.4 & 63.8$\pm$1.1 & 64.5$\pm$1.1  \\
\midrule
\end{tabular}
\end{adjustbox}
\end{table}

\begin{figure}[t]
\begin{center}
\includegraphics[width=\linewidth]{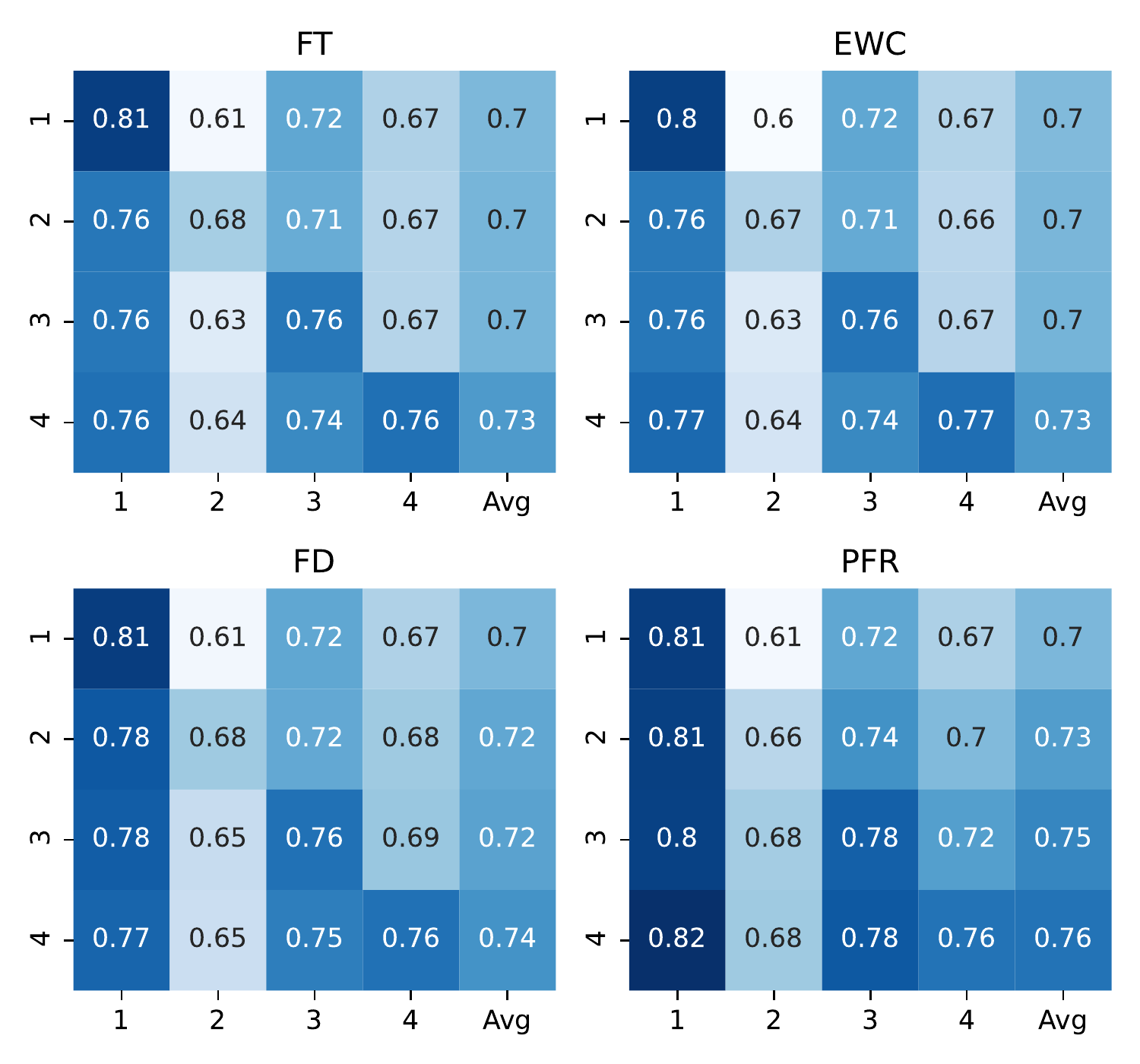}
\caption{Task-aware performance after the four consecutive tasks on CIFAR-100 for several methods.  Each row reports the results after each task, and columns represent on which task the model is evaluated. The last column reports average accuracy after each trained task.}
\label{fig:acc_matrix}
\end{center}
\end{figure}

In addition, we show the task-aware results on CIFAR 100 of the incrementally learned representations in Figure~\ref{fig:acc_matrix}. Here the models are the same as those used in Table~\ref{tab:cifar100_main}. Note that all other results in the paper are task-agnostic (no task-ID given during inference). Here, we also use training data from future tasks to train the classifier: this allows us to also evaluate the performance of the tasks that have not been seen by the feature extractor (see above diagonal elements).
We observe that all methods, including FT, incrementally improve results. 
Only our proposed PFR method considerably outperforms FT in this setting. It is also interesting to observe that PFR obtains positive backward transfer, since the performance on task 1 and 2 improves during the consecutive training sessions.

\begin{figure}[t]
\begin{center}
\includegraphics[trim=1.8cm 1.2cm 1.5cm 1.2cm, width=1\linewidth]{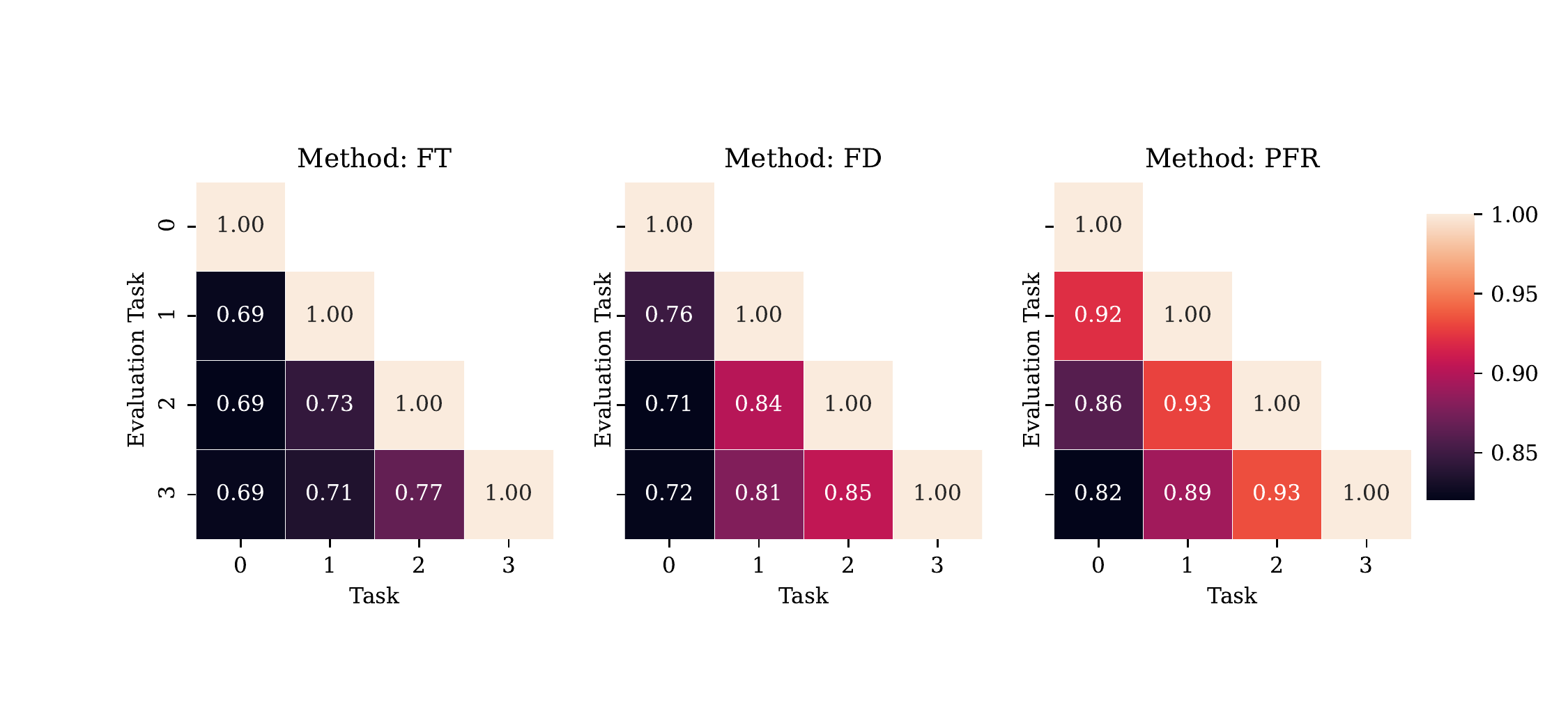}
\caption{Representation similarity compared with CKA for FT, FD, and PFR during incremental training.}
\label{fig:cka}
\end{center}
\end{figure}

\minisection{Learned representations.} In addition to evaluating accuracy on downstream classification tasks, we compare learned representation similarity with a Centered Kernel Alignment (CKA)~\cite{kornblith2019cka}. The results are given in Figure~\ref{fig:cka}. When the task is learned and immediately evaluated, the similarity is equal to one. When we finetune the model with new data, we start experiencing representation degradation -- seen in decreasing values in the columns in Figure~\ref{fig:cka}, left. With PFR, representation forgetting progresses much more slowly. FT is the worst, having the first task similarity after the last one with CKA equal $0.69$, next is FD with value $0.72$, and the best is PFR with $0.82$.

\begin{figure*}[t]
\begin{subfigure}[c]{0.36\textwidth}
    \includegraphics[width=\textwidth]{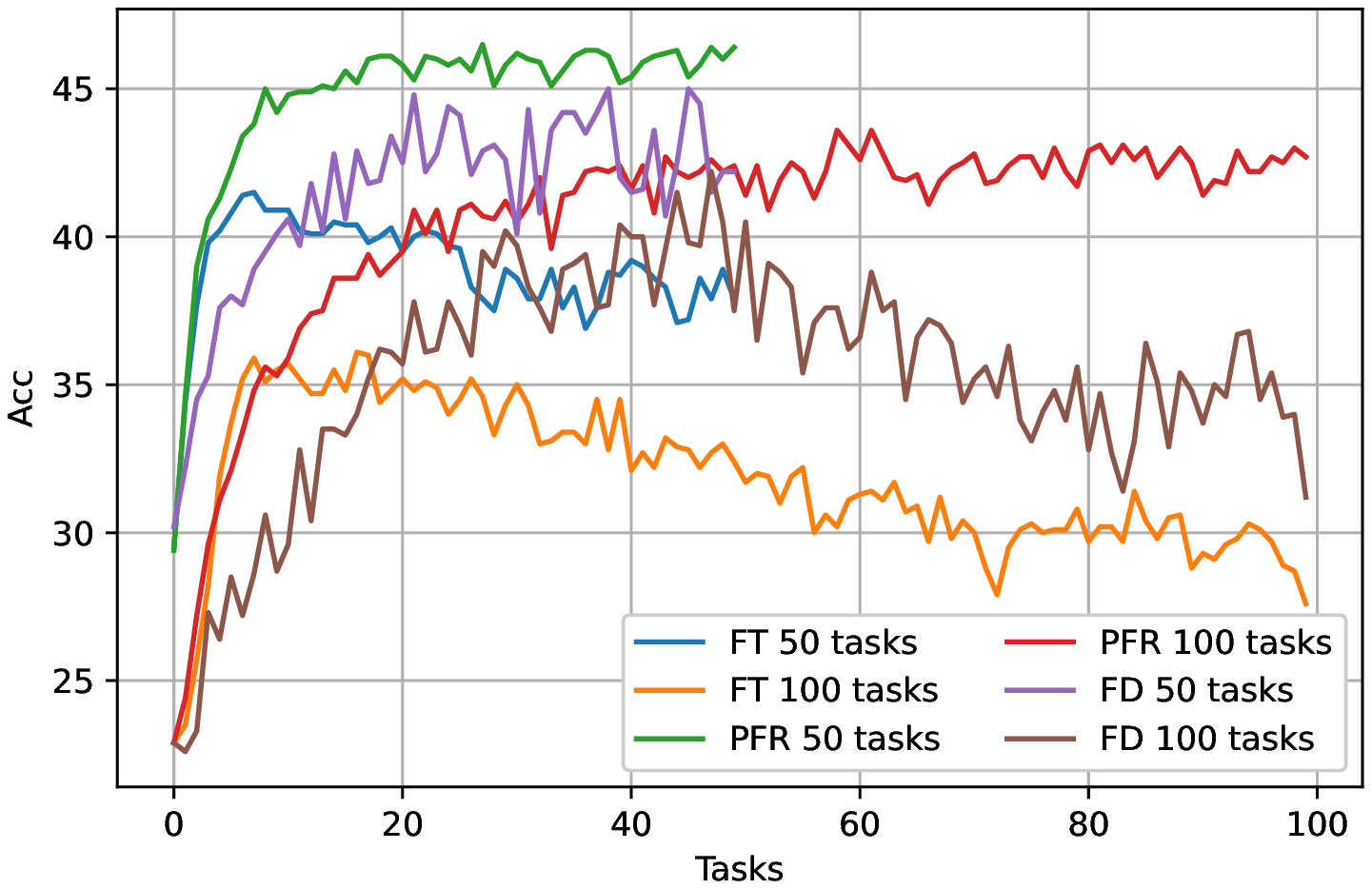}
    \caption{}
    \label{fig:more_task_scenario}
\end{subfigure}
\hfill
\begin{subfigure}[c]{0.3\textwidth}
    \includegraphics[width=\textwidth]{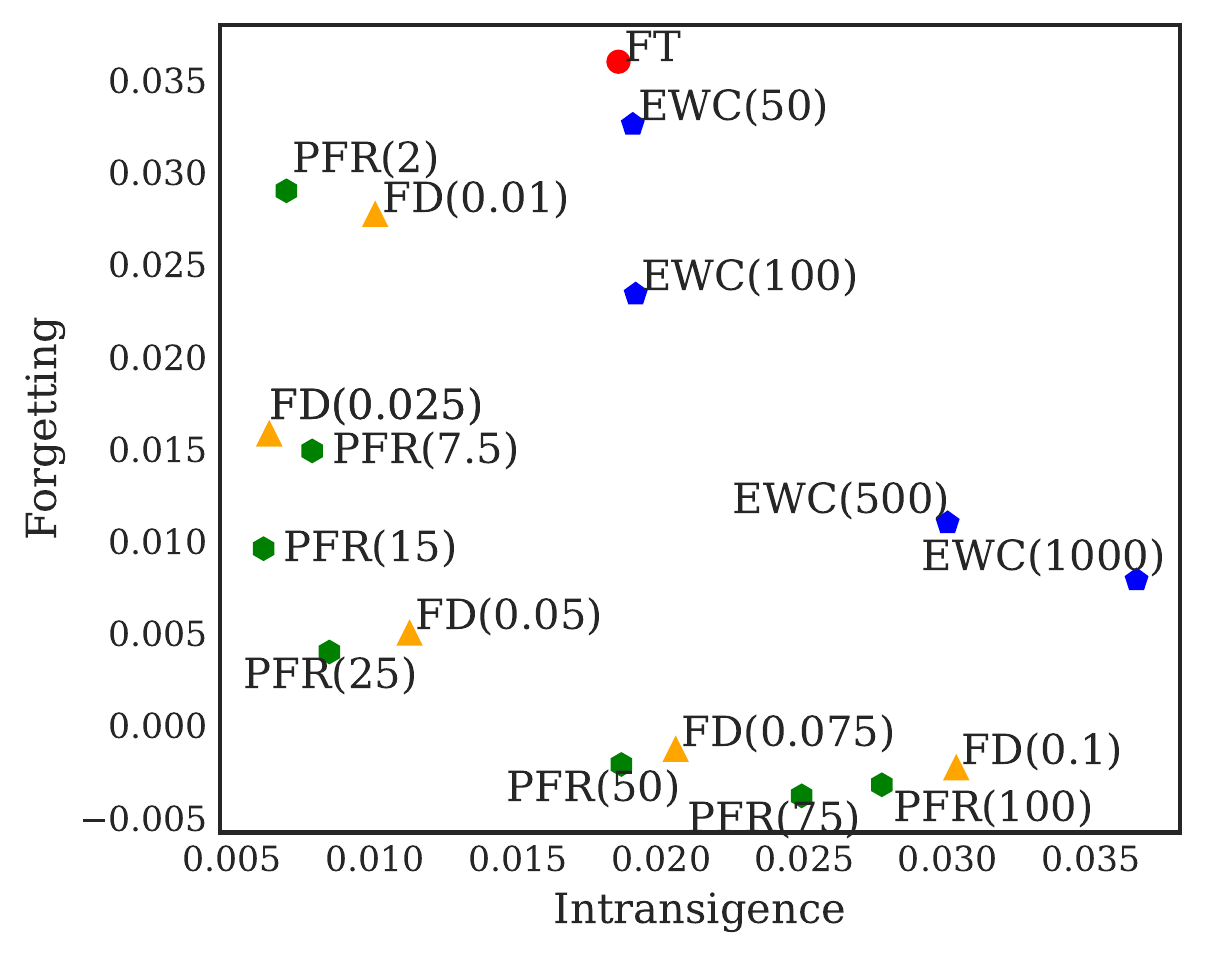}
    \caption{}
    \label{fig:reg}
\end{subfigure}
\hfill
\begin{subfigure}[c]{0.3\textwidth}
    \includegraphics[width=\textwidth]{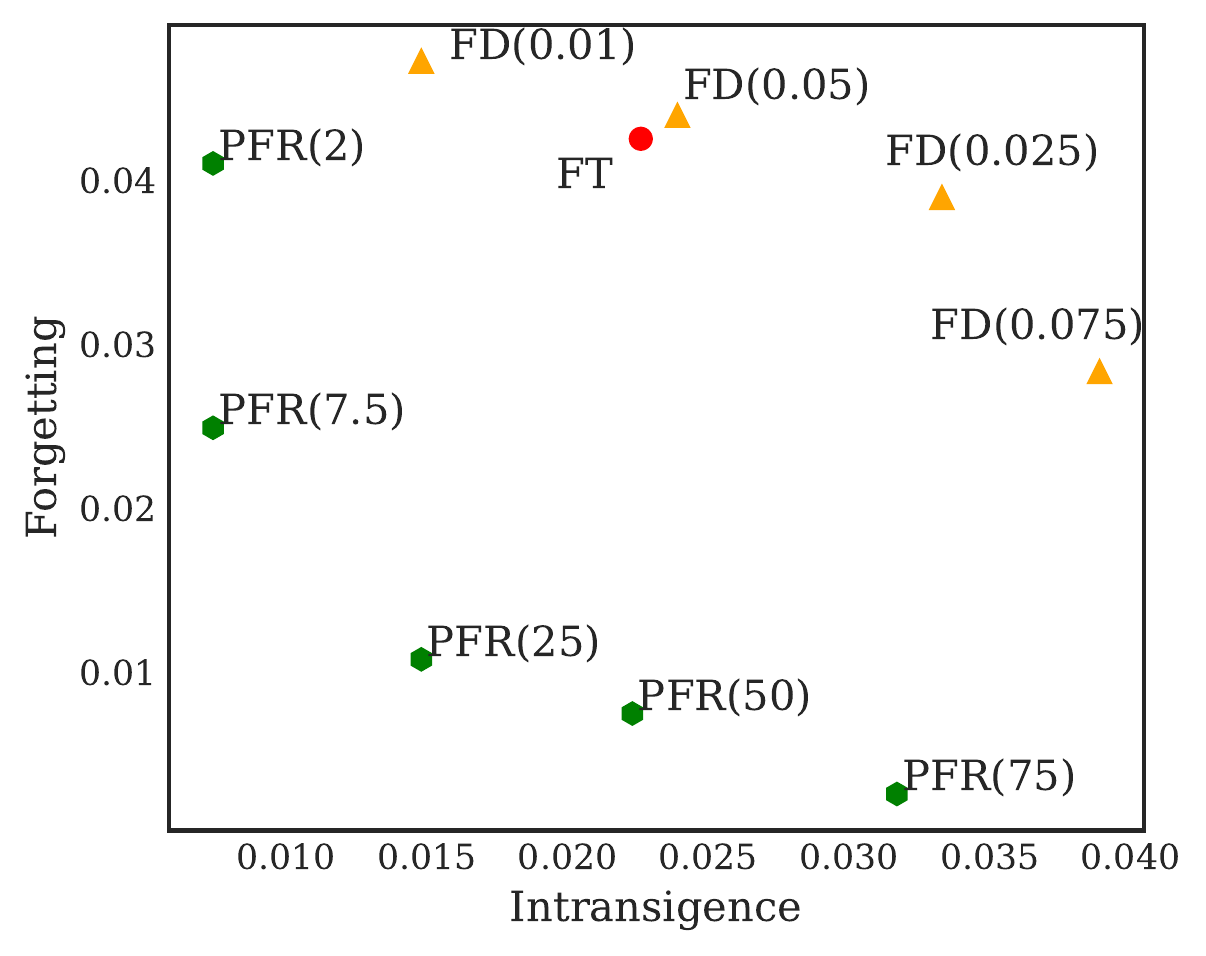}
    \caption{}
    \label{fig:reg_tiny}
\end{subfigure}

\caption{(a): Performance of several methods for different numbers of tasks on CIFAR-100. (b, c): Influence of regularization on forgetting and intransigence for EWC, FD, and PFR. FT uses no regularization and represents a point of reference. The regularization hyperparameter $\lambda$ is given in parentheses, (b) presents results for CIFAR-100 dataset, (c) for Tiny ImageNet.}
\end{figure*}

\minisection{Many task scenario.} Here we consider a challenging setting with longer sequences -- i.e. with more tasks. We experimented with our PFR method, fine-tuning(FT) and feature distillation(FD) on CIFAR-100 split into 50 or 100 tasks. In the case of 100 tasks, we only have a single class per task, 
which is an interesting setting since there are no negative classes, forcing the network to learn representations that are discriminative at the instance level. Results are presented in Figure~\ref{fig:more_task_scenario}, where accuracy over all classes is given per training session for each method. 
Without any mitigation of forgetting, FT cannot maintain the learned representations in longer tasks sequences, dropping closely to the level of a randomly initialized backbone (23\%) in the extreme case of 100 tasks. FD is also struggling on the longer sequence of 100 tasks. Only, our method shows stable results, preventing forgetting of the learned representation and progressing steadily. Similar results can be observed for 50 tasks, however the differences are not as pronounced as for the 100 task sequence.


\subsection{The influence of regularization}
Each regularization method is applied differently to the self-supervised network. To assess the influence of regularization and quantify its effect, we use the \emph{forgetting} and \emph{intransigence} measures defined in \cite{chaudhry2018riemannian}. Forgetting measures the average drop in accuracy per-task during continual learning. Intransigence describes the inability of a model to learn a new task. Formally, it is the difference between a referential model accuracy at task $t$ -- for us jointly trained self-supervised model up to task $t$ -- and the current task accuracy measured using held-out data. 

In Figure~\ref{fig:reg} all methods with different $\lambda$ parameter values are shown for CIFAR-100 dataset. FT is a point of reference here since it uses no regularization and maximum forgetting is expected. EWC is a weight regularization method that regularizes network weight changes using the Fisher Information Matrix. With larger lambda, forgetting is lower, but at the same time we pay the price of larger intransigence. As in the other experiments, we found that weight regularization does not obtain satisfactory results when applied to continual self-supervised learning. Feature distillation represents a better trade-off. The closer the results are to the bottom left corner, the better they are, and here PFR is the clear winner. The PFR results are consistently better than FD by some margin, which implies that the additional flexibility of the model introduced by the temporal projection network indeed leads to higher plasticity while at the same time keeping forgetting low. These results are confirmed on the larger Tiny ImageNet dataset (see Figure~\ref{fig:reg_tiny}). Here, the gap between FD and PFR is much larger, showing that projected functional regularization suffers from much lower forgetting at equal intransigence values.



\subsection{Generality of the Approach}

In order to verify that PFR generalizes to other self-supervised approaches, we conducted a series of experiments with SimCLR~\cite{chen2020SimCLR}, SimSiam~\cite{chen2021SimSiam}, and BarlowTwins~\cite{zbontar2021barlow}. In Table~\ref{tab:cifar100_SimCLR} we present results of fine-tuning and PFR for a ten task scenario. PFR  results in $5.4\%$, $8.4\%$, and $8.1\%$ improvement over FT after the final task. For this longer task sequence scenario the gain of our method with respect to FT is much larger when compared with results in  Table~\ref{tab:cifar100_main}. Starting from the second task, the effect of projected regularization is clear. In the ten task scenario, SimSiam after the first five tasks begins to outperform BarlowTwins, which is reflected by the results in the Task 10 column.

\begin{table}
\caption{Accuracy of SimCLR, SimSiam and BarlowTwins on CIFAR-100 split into 10 tasks of 10 classes.}
\label{tab:cifar100_SimCLR}
\centering
\begin{tabular}{l|rrrrrr}
\toprule
\textbf{Method} & \textbf{Task 1} &\textbf{Task 2} & \textbf{Task 5} & \textbf{Task 10} \\
\midrule\midrule
SimCLR PFR    &   40.4     & 44.7 & 47.2 & 48.2 \\
SimSiam PFR   &   43.1     & 50.2 & 53.1 & 55.1 \\
BarlowT PFR   &   45.1     & 50.6 & 54.7 & \textbf{55.4} \\
\midrule
SimCLR FT     &   40.4      & 41.2 & 40.6 & 42.8 \\
SimSiam FT    &   43.1       & 46.0 & 47.0 & 46.7 \\
BarlowT FT    &   45.1       & 48.8 & 48.9 & \textbf{47.0} \\

\bottomrule
\end{tabular}\label{tab:tin_others}
\end{table}

\subsection{Transfer to downstream tasks}

To better asses the quality of the trained representation, we evaluated all methods on a series of different downstream datasets. This allows us to evaluate the transferability of the learned features during the continual training process. The results for the smaller sized (32x32 images) CIFAR-100 and SVHN datasets are in Table~{~\ref{tab:cifar100_main} (bottom table)}. We observe a similar outcome as in the source dataset evaluation. The best results use our PFR method, followed by FD, EWC, and finally FT. The results are consistent during incremental learning: the better the representation is on a source classification task, the better it is on the target (SVHN) dataset after each task.

Furthermore, we trained all the methods on a larger dataset (Tiny ImageNet). We use the same procedure as for CIFAR-100 with four tasks, but with bigger images ($64 \times 64$) and more classes (200). The results are presented in Table~\ref{tab:tin_main}. In this dataset our method (PFR) surpasses FD, which is followed by the modified EWC and FT. 

As in CIFAR100 we evaluate our networks trained on Tiny Imagenet on different downstream datasets (Cars and Aircraft). The results are given in Table~\ref{tab:tiny_downstream}. For these datasets, as in CIFAR100, the accuracy of the various techniques follows the same pattern: PFR yields the best results, followed by FD, EWC, and finally FT.



\begin{table}
\caption{Accuracy on TinyImageNet split into four tasks. The learned representation is evaluated using a linear classifier over all classes after each task.}
\label{tab:tin_main}
\centering
\begin{tabular}{l|rrrrr}
\toprule
\textbf{Method} & \textbf{Task 1} & \textbf{Task 2} & \textbf{Task 3} & \textbf{Task 4}\\
\midrule\midrule
Joint (no CL)   &  - & - & - & 46.0 \\
\midrule
FD              & 35.3 & 35.9 & 36.6 & 36.6 \\
EWC             & 35.3 & 37.4 & 36.9 & 38.2\\
PFR (Ours)      & 35.3 & 38.6 &  39.9&\textbf{42.3} 
\\
\midrule
FT              & 35.3 & 38.3 & 38.5 & 39.1\\

\bottomrule
\end{tabular}
\end{table}

\begin{table}[]
\caption{Transfer Learning to downstream tasks.}
\label{tab:tiny_downstream}
\centering
\begin{tabular}{l|llll}
\midrule
\multicolumn{4}{c}{Cars}                \\
\midrule
\textbf{Method }     & \textbf{Task 1} & \textbf{Task 2} & \textbf{Task 3} & \textbf{Task 4}\\
\midrule\midrule
Joint (no CL) &   -     &   -    &   -   &   34.5    \\
\midrule
FD       &   27.1    & 30.6    & 31.8   &   31.1     \\
EWC       &  27.1     &  28.3    &  27.7   &   27.8  \\
PFR       &  27.1      & 31.1   &  33.1   &  \textbf{33.8}  \\
\midrule
FT          & 27.1      & 29.5    &  29.2    & 31.5      \\
\midrule
\multicolumn{4}{c}{Aircraft}            \\
\midrule
\textbf{Method }     & \textbf{Task 1} & \textbf{Task 2} & \textbf{Task 3} & \textbf{Task 4} \\
\midrule\midrule
Joint (no CL) &   -     &    -    &   -  &   30.0     \\
\midrule
FD       &  24.6     &  25.6  &  27.0    &  27.0     \\
EWC       &  24.6     & 23.8  & 25.3 &  25.1   \\
PFR       &  24.6     & 26.2  &  27.0   &  \textbf{28.4}  \\
\midrule
FT          & 24.6     & 25.6   &  26.9   & 27.0     \\
\midrule

\end{tabular}
\end{table}

\section{Conclusions and Future directions}~\label{sec:Conclusions}
In this paper, we proposed a method for incremental self-supervised learning without the need for any stored examples of previous tasks. Most existing regularization methods for continual learning are applied to class predictions or logits. Such approaches applied to self-supervised representation learning result in low plasticity. To address this, we propose Projected Functional Regularization via a temporal projection network that ensures that the newly learned feature space preserves information of the previous one, while still allowing for the learning of new features, resulting in higher plasticity. Extensive results on CIFAR100 and Tiny ImageNet demonstrate that our approach outperforms standard feature distillation by a considerable margin.

Finally, there are several limitations and future directions we discuss here. First, due to the high computational demands, experiments have been performed on low-resolution images, and they need to be confirmed for higher resolution data. Next, our method assumes access to task boundaries and cannot be directly applied in the task-free setting (without task boundaries)~\cite{aljundi2019task}. We think that this can be addressed by replacing the regularization based on the model from the previous task, with a regularization model that is updated with momentum. Next, continual learning of transformer architectures has only recently started~\cite{pelosin2022towards,Douillard2021}.  Self-supervised learning is a key ingredient of the transformer network training, and integrating our theory with these attention-based architectures for their continual learning is especially interesting. Finally, extending the theory with a limited replay buffer is of interest and would allow to directly report class-incremental learning results where the buffer is used to compute the classifier layer.

\section*{Acknowledgements}
We acknowledge the support from Huawei Kirin Solution, and the Spanish Gouvernement funded project PID2019-104174GB-I00/ AEI / 10.13039/501100011033. 

\newpage
{\small
\bibliographystyle{ieee_fullname}
\bibliography{main}
}

\end{document}